# USING A NEW PARSIMONIOUS AHP METHODOLOGY COMBINED WITH THE CHOQUET INTEGRAL: AN APPLICATION FOR EVALUATING SOCIAL HOUSING INITIATIVES

Francesca Abastante[1], Salvatore Corrente[2], Salvatore Greco[2,3], Alessio Ishizaka[3], Isabella M. Lami[1]

Abstract: We propose a development of the Analytic Hierarchy Process (AHP) permitting to use the methodology also in cases of decision problems with a very large number of alternatives evaluated with respect to several criteria. While the application of the original AHP method involves many pairwise comparisons between alternatives and criteria, our proposal is composed of three steps: (i) direct evaluation of the alternatives at hand on the considered criteria, (ii) selection of some reference evaluations; (iii) application of the original AHP method to reference evaluations; (iv) revision of the direct evaluation on the basis of the prioritization supplied by AHP on reference evaluations. The new proposal has been tested and validated in an experiment conducted on a sample of university students. The new methodology has been therefore applied to a real world problem involving the evaluation of 21 Social Housing initiatives sited in the Piedmont region (Italy). To take into account interaction between criteria, the Choquet integral preference model has been considered within a Non Additive Robust Ordinal Regression approach.

Keywords: Analytic Hierarchy Process, Parsimonious preference information, Social Housing, Choquet Integral, Robust Ordinal Regression

## 1. Introduction

Real world decision problems ask very often for methodological developments permitting to deal with their high complexity. When this happens, it is a very fortunate opportunity to advance both on the operational and the theoretical

[1] Department of Regional and Urban Studies and Planning (DIST), Politecnico di Torino, Viale Mattioli 39, 10125 Turin, Italy.

[2] Department of Economics and Business, University of Catania, Corso Italia 55, 95129, Catania, Italy.

[3] University of Portsmouth, Portsmouth Business School, Centre of Operations Research and Logistics (CORL), Richmond Building, Portland Street, Portsmouth.

Email addresses: francesca.abastante@polito.it (F. Abastante), salvatore.corrente@unict.it (S. Corrente), salgreco@unict.it (S. Greco), Alessio.Ishizaka@port.ac.uk (A. Ishizaka), isabella.lami@polito.it (I.M. Lami).

ground. Indeed, on one hand, the requirements emerging from the real world problem justify the methodological progress and prevent it from being relegated to a mere abstract level. Moreover, the successful application of the new methodology constitutes its best validation. On the other hand, the soundness of the theoretical basis of the new methodology contributes to improve the standard of the current practices in decision support.

The opportunity to progress on the methodological level starting from a real world problem is at the basis of this paper. Indeed we started our research from a real world problem acquiring more and more attention, that is the Social Housing (SH).

The SH projects are difficult to evaluate. The reason is mainly related to the fact that the SH concept has become increasingly complex from social, economic, financial, technical and environmental point of views. Moreover, while after the second world war the focal point of the SH was to provide houses to people in an emergency situation, over the last 20 years the human factor has become fundamental. The SH focus has shifted from the building to the people living in the building (CECODHAS Housing Europe, www.housingeurope.com). In this sense, the beneficiaries of the SH encompass people not having the minimum income needed to pay a rent on the regular real-estate market and people needing social support (Marx and Nolan 2012).

The aforementioned changes are mainly due to the economic crisis that affected Europe in recent years causing an erosion of public and private economic resources and contributing to the growth of new phenomena. These multifaceted circumstances call for integrative approaches in order to consider properly all those aspects overcoming the mere approach based on the economic and technical feasibility of the projects (Lami and Abastante 2017).

The new methodological approach we are proposing was applied to a real Italian case study that we had the opportunity to face thanks to the cooperation of the *Programma Housing*, which is an operating entity of the Italian Bank Foundation Compagnia di San Paolo located in Turin (Italy). The *Programma Housing* gives grant contributions to third bodies submitting innovative SH projects (from 2007, 6.150.000 euro have been given). The research presented in the paper addresses the issue of evaluating SH projects proposing a methodological approach that allows to tackle decision problems characterised by: high number of alternatives to rank, qualitative and quantitative criteria which could violate the preference independence, the possibility for the Decision Maker (DM) to express her preferences only on the alternatives she knows best.

Thus we imagined an approach permitting to organize the information by alternating stages of dialogue and calculation. The dialogue stages aim at collecting information directly from the DMs, which can reveal their preferences about the alternatives and the criteria at stake. The DMs preferences are in turn taken into account in the calculation stages.

The methodological approach proposed is based on the conjoint application of Analytic Hierarchy Process (AHP) and Choquet integral within Robust Ordinal Regression (Corrente et al. 2016). There, the basic idea was the transformation of an objective numerical evaluation of a considered criterion in a subjective measure of attractiveness permitting comparability between performances of all criteria through application of AHP on a small set of reference levels for each criterion and interpolation for all other levels. For example, in a decision regarding cars, the objective numerical evaluation of the maximum speed of a given model is transformed through this methodology in a subjective utility value representing the attractiveness for the DM. In the real world decision problem considered in this paper, the methodological challenge of this approach was related to the presence of criteria for which there was not a pre-existing objective numerical evaluation, so that the interpolation proposed in Corrente et al. 2016 would not be applied. Instead, according to the basic AHP methodology, it would be necessary to build the prioritization of each alternative on the basis of the pairwise comparisons of each alternative with all the others. This would require asking the experts involved in the decision problem to supply a huge quantity of information, that is a comparison with respect to strength of the preference for all couples of alternatives with respect to all considered criteria. In the considered decision problem, with 21 alternatives and 9 criteria without a pre-existing numerical evaluation (there was also a tenth criterion with an objective numerical scale on which the interpolation procedure could be applied) it would be necessary to ask 210 pairwise comparisons for each one of the non-numerical criteria for a total of 1,890 pairwise comparison! This is a huge amount of preference information. To give an idea of the effort asked to the DM, let us suppose that for each pairwise comparison, in average, the experts take one minute. Then they should take 31 hours and half to supply all these preferences. Realistically, this cannot be asked to the experts. This bottleneck is well known in the literature on AHP: for example, Saaty and Odzemir (2007) demonstrate that, using AHP, the number of elements to be considered should be no more than seven. Thus, what to do? Should we renounce to the use of AHP in case of a number of alternatives larger than seven? As the reader can imagine this is a central problem for the application of a very well known and appreciated method as AHP. Indeed, real world decision problems very often have more than seven alternatives, and, even more, the case in which the alternatives are seven or less is really a small subset of the family of decision problems we encounter in real life. Thus, to handle our decision problem, but more in general, to get a new methodology permitting to use AHP also in case the number of alternatives is larger than seven, we developed a new methodology permitting to continue to use AHP, but reducing the preference information asked to the DM (in our case, the experts involved in the SH decision problem). The new proposal is composed of four steps: in the first step, the DM has to provide a direct rating of the alternatives on the considered criteria; in the second step the DM, in accordance with the analyst, has to select some reference evaluations on each criterion; in the third step, the DM has to compare the reference evaluations so that their prioritization is obtained by using the original version of the AHP method; in the fourth step, all the other evaluations are prioritized by interpolation according to the priority values obtained for the reference evaluations. The new proposal reduces in a considerable way the enormous cognitive effort asked to the DM in comparing alternatives and criteria in problems of big dimensions. However, it maintains the basic idea of AHP of relying on pairwise comparisons to prioritize the elements to be evaluated. Thus, beyond our specific decision problem, this is a procedure that can be applied in any case the application of original AHP is prevented by a number of alternatives larger than seven.

In order to validate the new proposal, we conducted an experiment with around 100 students of the University of Catania (Italy) that were requested to evaluate the area of ten geometric figures. The students were split in two groups of similar size. All of them managed the same problem even if the components of one group applied the original AHP method, while the other ones used the methodology we are proposing. The results showed that the new proposal gives better results than those obtained using the original methodology. Beyond this experimental validation of the proposed procedure, we had also a practical validation given by the appreciation expressed by the experts involved in the SH decision problem. In fact, they were able to supply the information we asked them, understanding the use of this information and acknowledging the interest of the recommendation supplied by the decision aiding procedure.

The paper is organised as follow. In the next section, we recall the basic concepts of the Choquet integral, Non Additive Robust Ordinal Regression (NAROR) and AHP, with a specific focus on how to use the latter for the definition of the evaluation scales and the comparison of several alternatives. Section 2 illustrates the methodological framework. Section 3 describes the application of the methodology to the case study; the discussion of the results is in section 4. We end the paper by offering some conclusions, highlighting the strengths and weaknesses of the proposed approach, and collecting further direction of research.

## 2. Methodological framework

### 2.1. Introduction on MCDA

Multiple Criteria Decision Aiding (MCDA; see Ishizaka and Nemery 2013 for a general introduction and Greco et al. 2016 for an updated and comprehensive collection of state of the art surveys), considers a set *A={a,b,c,…}* of alternatives evaluated with respect to a coherent family of points of view, technically called criteria, $G = \{g_1, g_2, \ldots, g_n\}$, with $g_i: A \rightarrow \mathbf{R}$, $i = 1, \ldots, n,$ such that, without loss of generality, we can suppose that for all $a, b \in A$ $g_i(a) \geq g_i(b)$ means that $a$ is at least as good as $b$ with respect to $g_i$. Four different decision problems can be considered with the methodologies developed within MCDA: ranking, choice, sorting and description (Roy, 1996). Since the only objective information stemming from the evaluations of the alternatives on the considered criteria, which is the dominance relation[2], is quite poor, different aggregation methods have been proposed in literature based on: value functions (Keeney and Raiffa 1976), binary relations (Brans and Vincke 1985, Roy 1996) or decision rules (Greco et al. 2001). Value functions assign to each alternative a real value being representative of the goodness of the considered alternative with respect to the problem at hand. Binary relations are the basis of the outranking methods for which $aSb$ means that $a$ is at least as good as $b$. Decision rules connect the evaluations of the alternatives on the considered criteria with the recommendations on the problem at hand. For example, "If the maximum speed of a car is at least 140 km/h and its consumption is at least 17 km/l, then the car is considered at least good".

Most of the methods known in MCDA assume that the set of criteria is mutually preferentially independent (Keeney and Raiffa 1976) even if it is evident that in real applications it is not always true. Indeed, sometimes, criteria present a certain

---

[2] $a$ dominates $b$ iff $a$ is at least as good as $b$ on all criteria and strictly better on at least one criterion

degree of negative or positive interaction. On one hand, criteria $g_i$ and $g_j$ are negatively interacting if the importance assigned to this couple of criteria is lower than the sum of the importance assigned to the two criteria singularly. On the other hand, $g_i$ and $g_j$ are positively interacting if the importance assigned to this couple of criteria is greater than the sum of the importance assigned to the two criteria singularly. In order to deal with such a type of interactions, non-additive integrals are used in literature and, among them, the most well-known are the Choquet integral (Choquet 1953; see Grabisch 1996 for a survey on the application of the Choquet integral in MCDA) and the Sugeno integral (Sugeno 1974).

## 2.2. Choquet integral

The Choquet integral can be considered as an extension of the known weighted sum aggregation method. It is based on a *capacity* (non-additive measure) being a set function $\mu: 2^G \rightarrow [0,1]$ such that the monotonicity constraints ($\mu(S) \leq \mu(T)$ for all $S \subseteq T \subseteq G$) and the normalization constraints ($\mu(\emptyset) = 0$ and $\mu(G) = 1$) are satisfied. Given a capacity $\mu$ and an alternative $a \in A$, the Choquet integral of the evaluations vector $g(a) = (g_1(a), \dots, g_n(a))$ with respect to $\mu$ is computed as follows:

$$C_\mu(a) = \sum_{j=1}^{n} \left[ g_{(j)}(a) - g_{(j-1)}(a) \right] \cdot \mu(N_i) \qquad (1)$$

where $(\cdot)$ is a permutation of the criteria indices such that $0 = g_{(0)}(a) \leq g_{(1)}(a) \leq \cdots \leq g_n(a)$ and $N_i = \{g_{(i)}, \dots, g_{(n)}\}$. To make calculations easier, a Möbius transformation of the capacity $\mu$ can be considered (Rota 1964, Shafer 1976), being another set function $m: 2^{|G|} \rightarrow [0,1]$ such that $\mu(S) = \sum_{B \subseteq S} m(B)$ for all $S \subseteq G$. Considering the Möbius transformation of the capacity $\mu$, the Choquet integral can be reformulated as follows

$$C_\mu(a) = \sum_{T \subseteq G} m(T) \cdot \min_{g_j \in T} g_j(a) \qquad (2)$$

while the monotonicity and normalization constraints are written as follows:

$1a)$ $m(\{g_i\}) + \sum_{T \subseteq R} m(T \cup \{g_i\}) \geq 0$ for all $g_i \in G$ and for all $R \subseteq G \setminus \{g_i\}$,

$2a)$ $m(\emptyset) = 0$ and $\sum_{T \subseteq G} m(T) = 1$.

The application of the Choquet integral involves the knowledge of $2^{|G|} - 2$ parameters (since $\mu(\emptyset) = 0$ and $\mu(G) = 1$), one for each subset of criteria in $G$. As the inference of all these parameters is quite complex, in general, only $k$-additive capacities are used. A capacity is said $k$-additive if $m(T) = 0$ for all $T \subseteq G$ having more than $k$ criteria. In the applications, 2-additive capacities are enough to represent the preferences of the DM. For this reason, in the following, we shall consider only 2-additive capacities. In this case, the Choquet integral of the evaluations vector $g(a) = (g_1(a), \dots, g_n(a))$ takes the form

$$C_\mu(a) = \sum_{j=1}^{n} m(\{g_j\}) \cdot g_j(a) + \sum_{\{g_i, g_j\} \subseteq G} m(\{g_i, g_j\}) \cdot \min\{g_i(a), g_j(a)\}$$

while the monotonicity and normalization constraints are the following:

$1b)$
$$\begin{cases} m(\{g_i\}) \geq 0, \quad \text{for all } g_i \in G, \\ m(\{g_i\}) + \sum_{g_j \in T} m(\{g_i, g_j\}) \geq 0, \text{ for all } g_i \in G \text{ and for all } T \subseteq G \setminus \{g_i\}, T \neq \emptyset, \end{cases}$$

$2b)$ $m(\emptyset) = 0$ and $\sum_{j=1}^{n} m(\{g_j\}) + \sum_{\{g_i, g_j\} \subseteq G} m(\{g_i, g_j\}) = 1$.

In this context, the importance of a criterion $g_i$ does not depend on itself only but also on its contribution to all coalitions of criteria. Consequently, the Shapley (Shapley 1953) and the interaction indices (Murofushi and Soneda 1993) can be defined as follows[3]:

$$\varphi(\{g_i\}) = m(\{g_i\}) + \sum_{g_j \in G \setminus \{g_i\}} \frac{m(\{g_i, g_j\})}{2},$$

$$\varphi(\{g_i, g_j\}) = m(\{g_i, g_j\}).$$

Despite its diffusion in MCDA, the use of the Choquet integral presents two main drawbacks:

1) As already underlined above, $2^{|G|} - 2$ parameters need to be known. The elicitation of these parameters can be done in a direct or in an indirect way. In the direct way, the DM is asked to provide directly values for all parameters involved in the model. In the indirect way, the DM provides some exemplary statements in terms of comparisons between alternatives or criteria from which parameters compatible with these statements can be found.

---

[3] Both the Shapley and the Murofushi indices can be defined for any capacity. Their definition is not restricted to 2-additive capacities only. For the formulation of these indices in the general case see, for example, Angilella et al. 2016.

While the direct technique is high cognitively demanding and the DM may be asked to provide parameters on which he has no expertise, in the indirect one the DM is more confident about her answer since she is providing opinions only on cases she knows very well.

2) In order to apply eqs. (1) or (2), the evaluations of the alternatives on the considered criteria have to be expressed on the same scale. In general this is not always the case and, consequently, a procedure aiming to build a common scale has to be used.

The AHP-Choquet method (Corrente et al. 2016), deals simultaneously with the two mentioned drawbacks. On one hand, the AHP (Saaty 1980, Saaty 1990) is applied to build a common scale for the alternatives' evaluations, while, on the other hand, the NAROR (Angilella et al. 2010a) takes into account the whole set of parameters compatible with the preferences provided by the DM in an indirect way assuming that the underlying preference model is the Choquet integral.

## 2.3. The Analytic Hierarchy Process

AHP is an MCDA method based on ratio scales for measuring performances on considered criteria and the importance of these criteria. The problem at hand is structured by AHP in a hierarchical way where the overall goal is set at the top of the hierarchy, the alternatives being the object of the decision are placed at the bottom of the hierarchy. The criteria on which the alternatives need to be evaluated are in the middle of the hierarchy between the overall goal and the alternatives themselves. Given $n$ criteria, the DM is supported to provide a value for evaluations $g_j(a), g_j \in G$ and $a \in A$ and for the weights $w_1, \ldots, w_n$ in a weighted sum aggregation. The basic idea is that it is more convenient to perform pairwise judgments rather than to give direct evaluations of performances $g_j(a)$ and weights $w_1, \ldots, w_n$ as experimentally seen in Millet 1997 and Por and Budescu 2017. For this reason, using AHP, for each criterion $g_j \in G$, the DM is asked to compare each couple of alternatives $\{a_r, a_s\}$ indicating the preferred alternative and expressing the degree of preference with a verbal judgement on a nine point scale

- 1: indifferent,
- 3: moderately preferred,
- 5: strongly preferred,
- 7: very strongly preferred,
- 9: extremely preferred

with 2, 4, 6 and 8 intermediate values between the two adjacent judgments. Denoting by $a_{rs}^{(j)}$ the pairwise comparison between the evaluations $e_r^{(j)}$ and $e_s^{(j)}$, it is possible to build a positive square reciprocal matrix $M^{(j)}$ of order $|A|$.

$$M^{(j)} = \begin{pmatrix} 1 & a_{12}^{(j)} & \cdots & a_{1|A|}^{(j)} \\ 1/a_{12}^{(j)} & 1 & \dots & a_{2|A|}^{(j)} \\ \vdots & \vdots & \ddots & \vdots \\ 1/a_{1|A|}^{(j)} & 1/a_{2|A|}^{(j)} & \cdots & 1 \end{pmatrix}$$

Several procedures have been proposed to determine the evaluations $e_r^{(j)}$ and among them the most well-known is that one based on the computation of the right eigenvector of the $M^{(j)}$ matrix (Saaty 1977), the row arithmetic mean vector, that is

$$e_r^{(j)} = \frac{\sum_{s=1}^{|A|} a_{rs}^{(j)}}{|A|}$$

and the row geometric mean vector (Crawford and Williams 1985), that is

$$e_r^{(j)} = \sqrt[|A|]{\prod_{s=1}^{|A|} a_{rs}^{(j)}}$$

If the entries of the matrix would be perfectly coherent, the following conditions should be satisfied

$$a_{rs}^{(j)} a_{sk}^{(j)} = a_{rk}^{(j)}$$

and the right eigenvector, the row arithmetic mean and the row geometric mean should give the same evaluations $e_r^{(j)}, r = 1, \dots, |A|$. As this is not the case in real world applications, the consistency of the judgments supplied by the DM can be tested computing the consistency ratio (CR):

CR=CI/RI

where $CI = (\lambda_{max} - |A|)/(|A| - 1)$ is the consistency index, $\lambda_{max}$ is the principal eigenvalue of $A$ and $RI$ is the ratio index. The ratio index is the average of the consistency indices of 500 randomly filled matrices. Saaty (1977) considers that a consistency ratio exceeding 10% may indicate a set of judgments too inconsistent to be reliable and therefore recommends revising the evaluations.

The same pairwise procedure is applied on the weights to calculate the local priorities $w_1, \ldots, w_n$. Once the priorities have been obtained, the final evaluation $U(a)$ of the alternative $a \in A$ is obtained as

$$U(a) = \sum_{j=1}^{n} e_a^{(j)} w_j.$$

## 2.4. Reduction of pairwise comparisons in AHP: the new proposal

As observed in the previous section, the application of the AHP method involves $\binom{n}{2} + n\binom{|A|}{2}$ pairwise comparisons ($\binom{n}{2}$ pairwise comparisons between importance of criteria and $\binom{|A|}{2}$ pairwise comparisons between alternatives on each considered criterion). The amount of information asked to the DM can be quite huge even for small problems. For example, considering a problem of reasonable dimension composed of 5 criteria and 10 alternatives, the DM has to provide $\binom{10}{2} + 10\binom{5}{2}$ pairwise comparisons, which are 145 pairwise comparisons. Consequently, it is quite difficult for the DM providing all these preferences because of her limited capacity of processing information (Miller 1956). If the criterion has an objective numerical evaluation (for example the maximum speed in a decision about cars), Corrente et al. 2016 proposed to apply AHP to prioritize a small set of reference levels of the considered criterion and to build the subjective measure of attractiveness of all the evaluations on the considered criterion by linear interpolation. Here we extend that approach to the case in which the considered criterion has not an objective numerical evaluation, so that the above mentioned interpolation cannot be applied and the basic AHP should be applied asking to the DM the pairwise comparisons of each alternative with all the other alternatives. Our proposal aims to reduce the cognitive effort involved in the application of the basic AHP and can be summarized in the following procedure:

Step 1- For each criterion $g_j$, we ask the DM to give a rating to alternatives in $A$ on a common scale (for example 0-100). We shall denote by $r_j(a)$ the rating provided by the DM to the alternative $a$ with respect to criterion $g_j$;

Step 2- For each criterion $g_j$, the DM, in accordance with the analyst, has to fix $t_j$ reference evaluations $\left(\gamma_{j1}, \dots, \gamma_{jt_j}\right)$ on the considered common scale;

Step 3- Following Corrente et al. 2016, the DM is therefore asked to apply the AHP to the set composed of the reference evaluations defined on Step 2 obtaining the normalized evaluations $u(\gamma_{js})$, for all $j = 1, \dots, n$ and for all $s = 1, \dots, t_j$;

Step 4- The rating of the evaluations provided by the DM which are not reference evaluations are obtained by interpolating the normalized evaluations got in the previous step. For each $r_j(a) \in \left[\gamma_{js}, \gamma_{js+1}\right]$, the following value is computed:

$$u\left(r_j(a)\right) = u(\gamma_{js}) + \frac{u(\gamma_{js+1}) - u(\gamma_{js})}{\gamma_{js+1} - \gamma_{js}} \left(r_j(a) - \gamma_{js}\right). \qquad (1)$$

While in the original AHP method, the DM was asked to provide the pairwise comparison of all pairs of alternatives on all considered criteria, in this case she is asked to provide, at first, the rating of the alternatives on the considered criteria and applying the AHP on a small subsets of reference evaluations defined for each criterion. Observe that if the considered criterion has an objective numerical evaluation the above procedure can be applied substituting this numerical evaluation to the direct rating supplied by the DM in Step 1. In fact, this is the procedure proposed in Corrente et al. 2016 for criteria with objective numerical evaluations.

## 2.5 An experiment on the reliability of the proposed approach

Before applying the methodology of scale construction presented in the previous section to the real world decision problem that we shall present in section 3, we have tested it through one experiment. Here is the description of the experiment and its results. Let us to point out that analogous experiments have been performed and described in Ishizaka and Nguyen Nam (2013) and Meesariganda and Ishizaka (2017).

We selected a group of 98 undergraduate students from the Department of Economics and Business of the University of Catania. This group was split in two subgroups of 46 and 52 students. Both groups were presented with a sheet on which ten geometrical figures (triangles, rectangles, circles, etc.) of different area were pictured (see Figure 1). Anyway, the components of the two groups were asked to fulfil two different assignments:

1.  Students belonging to the first group were asked to:
    - give an estimate of the area of the geometric figures B-L, knowing that the area of the geometric figure A was equal to 1,
    - fill in a pairwise comparison matrix related to the five geometric figures shown in figure 2, giving an evaluation of the ratio between their area using the usual 9-point qualitative scale used by AHP (in this case "equally large", "weakly larger", "strongly larger", "very strongly larger", "absolutely larger" corresponding to 1, 3, 5, 7 and 9, respectively, with 2, 4, 6, 8 being related intermediate evaluations).

2.  Students belonging to the second group were asked to fill in the pairwise comparison matrix relative to all ten figures as in the original AHP method. This assignment involved $\binom{10}{2} = 45$ pairwise comparisons of the different areas.

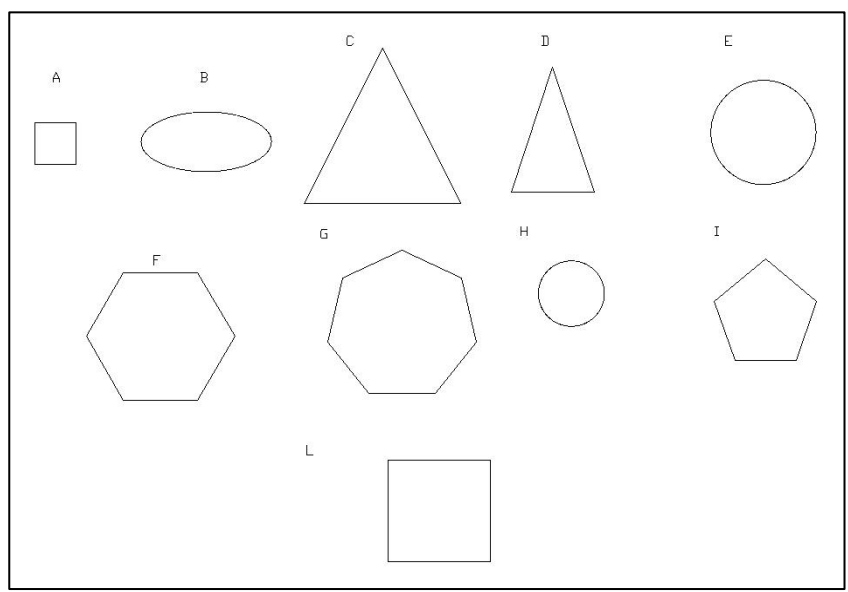

**Figure 1.** Geometrical figures which area were estimated by both groups of students

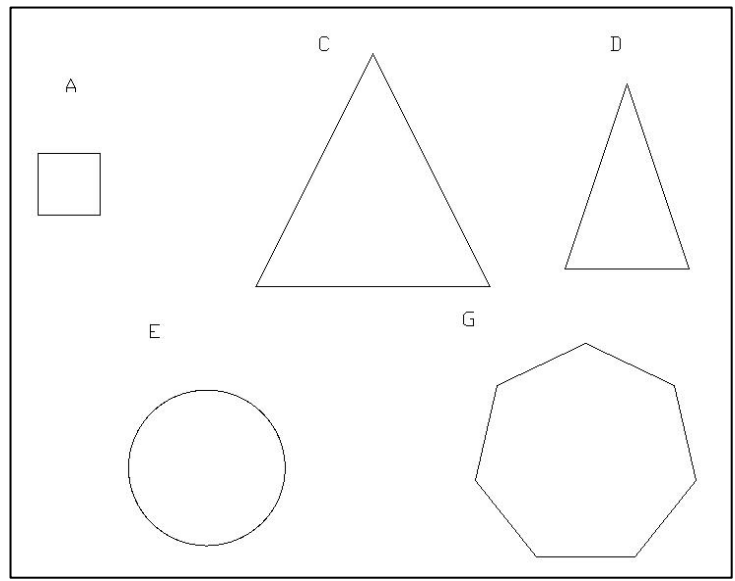

**Figure 2.** Geometrical figures pairwise compared to correct the direct estimation

For the students of the first group we computed the area of the ten geometric figures according to the methodology described in section 2.4[4], while for the students of the second group we computed their area using the usual AHP approach, that is, by computing the normalized right principal eigenvectors of the comparison matrices. Finally, denoting by $Area^{real}$ and $Area^{estimated}$ the vectors containing the real area and the estimated area of the ten geometrical figures shown in Figure 1, respectively, we computed the mean squared error (MSE) of these two vectors, that is,

$$MSE = \frac{1}{10}\sum_{i=1}^{10}\left(Area_i^{real} - Area_i^{estimated}\right)^2 .$$

---

[4] Note that, in this case, the set of reference evaluations mentioned in step 3 of the procedure described in section 2.4 is the set of geometric figures shown in Figure 2. Therefore, students belonging to the second group had to perform $\binom{5}{2} = 10$ pairwise comparisons only.

Let us observe that by using the procedure described in section 2.4 two different area vectors will be considered:

- the vector composed of the area estimated by the students before applying the AHP method (this is what we shall call "direct estimation in Table 1 below"), and
- the vector composed of the area of the ten geometric figures estimated with the application of the whole procedure (this is what we call "Direct estimation with correction" in Table 1 below).

Moreover, we stored the time necessary to each student to complete the assigned task. We computed the mean and standard deviation of MSE and the mean and standard deviation of the time taken by each student to complete her assignment. To clean the data from estimations not enough careful, we considered the consistency ratio of the provided pairwise matrices, computing again the mean and the standard deviation of the MSE only for students whose corresponding pairwise comparison matrices had a CR smaller than 10%. In this way we removed 12 questionnaires from the first group of students and 6 questionnaires from the second group of students. All these data are shown in Table 1.

**Table 1.** Results of the experiment

|  | ALL QUESTIONNAIRES | | | QUESTIONNAIRES WITH CR LOWER THAN 10% | | |
|---|---|---|---|---|---|---|
|  | Original AHP | Direct estimation | Direct estimation with correction | Original AHP | Direct estimation | Direct estimation with correction |
| Average MSE | 11.2780 | 6.800744 | 10.17001 | 10.7239 | 6.8864 | 6.6076 |
| Std MSE | 7.6621 | 21.4968 | 10.6047 | 6.8388 | 24.0518 | 4.8062 |
| Average time | 18.285714 | 16.42222 | 16.42222 | 18.75 | 16.4 | 16.4 |
| Std time | 3.6512 | 5.0113 | 5.0113 | 3.8326 | 4.9539 | 4.9539 |

The data in Table 1 show that considering all questionnaires, the best results, in terms of MSE, are obtained by the direct estimation. However, let us observe that there is a great standard deviation that suggests a great instability of the method. From this point of view, AHP is the most stable procedure, but its average MSE is greater than both the average MSE of the direct estimation and that one the directed estimation with correction. If, as suggested by Saaty, we consider only the pairwise comparison matrices with a CR lower than 10%, the best method is clearly the direct estimation with correction that has both the smallest MSE and the smallest standard deviation. The results of this experiment encouraged us to apply the method of direct estimation with correction using AHP to the real world problem of SH evaluation.

## 2.6. Non Additive Robust Ordinal Regression

Once the common scale for the evaluations of the alternatives on the considered criteria has been built, as already observed in section 2.2, one has to provide the capacity needed for the application of the Choquet integral preference model and, in general, the indirect technique is preferred by the DM. In this case, the DM is asked to provide some preferences in terms of comparisons between alternatives ($a$ is preferred to $b$ ($a \succ b$) or $a$ is indifferent to $b$ ($a \sim b$), or $a$ is preferred to $b$ more than $c$ is preferred to $d$ ($(a,b) \succ^* (c,d)$) etc.); or in terms of comparisons between criteria ($g_i$ is more important than $g_j$ ($g_i \succ g_j$), or $g_i$ and $g_j$ are equally important ($g_i \sim g_j$) etc.) and interaction between criteria ($g_i$ and $g_j$ are positively interacting or $g_i$ and $g_j$ are negatively interacting, etc.). Once these preference statements are translated to constraints representing them, in order to check if there exists at least one compatible measure, one has to solve the following LP problem[5]:

$$\max \quad \varepsilon \text{ s.t.}$$

$$\left.\begin{aligned}
&C_\mu(a\ ) \geq C_\mu(b\ ) \text{ if } a\ \succsim b \\
&C_\mu(a\ ) \geq C_\mu(b\ ) + \varepsilon \text{ if } a\ \succ b\ , \\
&C_\mu(a\ ) = C_\mu(b\ ) \text{ if } a\ \sim b\ , \\
&C_\mu(a\ ) - C_\mu(b\ ) \geq C_\mu(c\ ) - C_\mu(d\ ) + \varepsilon \text{ if } (a\ ,b\ ) \succ^* (c\ ,d\ ), \\
&C_\mu(a\ ) - C_\mu(b\ ) = C_\mu(c\ ) - C_\mu(d\ ) \text{ if } (a\ ,b\ ) \sim^* (c\ ,d\ ), \\
&\varphi(\{g_i\}) \geq \varphi(\{g_j\}) + \varepsilon \text{ if } g_i \succ\ g_j, \\
&\varphi(\{g_i\}) = \varphi(\{g_j\}) \text{ if } g_i \sim\ g_j, \\
&\varphi(\{g_i, g_j\}) \geq \varepsilon, \text{ if criteria } g_i \text{ and } g_j \text{ are positively interacting,} \\
&\varphi(\{g_i, g_j\}) \leq -\varepsilon, \text{ if criteria } g_i \text{ and } g_j \text{ are negatively interacting,} \\
&m(\{\emptyset\}) = 0, \sum_{g_i \in G} m(\{g_i\}) + \sum_{\{g_i, g_j\} \subseteq G} m(\{g_i, g_j\}) = 1, \\
&m(\{g_i\}) \geq 0, \forall g_i \in G, \\
&m(\{g_i\}) + \sum_{g_j \in T} m(\{g_i, g_j\}) \geq 0, \forall g_i \in G, and\ \forall T \subseteq G \backslash \{g_i\}, T \neq \emptyset.
\end{aligned}\right\} E$$

---

[5] Let us observe that an auxiliary variable $\varepsilon$ in $E$ is used to transform strict inequalities into weak inequalities. For example, the constraint $C_\mu(a) > C_\mu(b)$, translating the strict preference of $a$ over $b$, is transformed into the constraint $C_\mu(a) \geq C_\mu(b) + \varepsilon$.

If $E$ is feasible and $\varepsilon^* > 0$, where $\varepsilon^* = \max \ \varepsilon$ s.t. $E$, then there exists at least one capacity compatible with the preference information provided by the DM. If this is not the case, then one has to check which constraints cause the infeasibility of $E$, by using some of the methodologies proposed in Mousseau et al. 2003. In this case, more than one capacity could be compatible with this preferences and the choice of only one of them could be arbitrary. For this reason, Robust Ordinal Regression (ROR; see Greco et al. 2008 for the paper introducing ROR and Corrente et al. 2013 for a survey on ROR) takes into account simultaneously all the capacities compatible with the preferences provided by the DM by defining a necessary and a possible preference relation: $a$ is necessarily preferred to $b$ if $a$ is at least as good as $b$ for all compatible capacities, while $a$ is possibly preferred to $b$ if $a$ is at least as good as $b$ for at least one compatible capacity. Considering the two sets of constraints

$$\left.\begin{array}{l} C_\mu(b) \geq C_\mu(a) + \varepsilon \\ E \end{array}\right\} E^N(a,b) \qquad \left.\begin{array}{l} C_\mu(a) \geq C_\mu(b) \\ E \end{array}\right\} E^P(a,b)$$

$a$ is necessarily preferred to $b$ if $E^N$ is infeasible or $\varepsilon^N \leq 0$, where $\varepsilon^N = \max \ \varepsilon$ s.t. $E^N$, while $a$ is possibly preferred to $b$ if $E^P$ is feasible and $\varepsilon^P > 0$, where $\varepsilon^P = \max \ \varepsilon$ s.t. $E^P$.

Once one gets the necessary and possible preference relations, the most representative value function can be computed (Angilella et al. 2010b). It is a value function compatible with the preferences provided by the DM summarizing the results got from the ROR since, on one hand, it maximizes the difference between alternatives $a$ and $b$ such that $a$ is strictly necessarily preferred to $b$, while, on the other hand, it minimizes the difference between alternatives $a$ and $b$ such that neither $a$ is necessarily preferred to $b$ nor the *vice-versa*. Details on the computation of the most representative value function can be found in Angilella et al. 2010b or Corrente et al. 2016.

### 3. The case study

In this section, the proposed methodological approach is illustrated with a Social Housing (SH) projects ranking. The issue of SH is currently challenging all over Europe and is characterized by specific peculiarities.

The houses destroyed during the conflict characterized the housing issue that affected Europe in the second post-war period. In this sense this phenomenon could be identified as "housing deprivation". By contrast, the current housing crisis does not derive from a shortage of available properties but is the result of a social, economic and cultural change.

The distinctive features of this crisis affect a wide segment of the population, also involving the middle classes. In fact, the so-called "grey zone", also known as "in-work poverty population", is composed of subjects in a situation of housing vulnerability or who need transitory housing solutions. The subjects belonging to the "grey zone" are identified as the people who cannot access to the real estate market but at the same time are not eligible to access the public housing programs: hidden homeless, immigrants, internal migrants, city users, single-earner families, the elderly, people subject to eviction, single parents, former prisoners.

The new growing housing demand is therefore characterized by high economic and social fragility (Wills and Linneker 2014). The rebalancing of the relationship between the number of households and the number of inhabitants is not dealt with but rather an attempt is made to lessen the gap between access to the housing market and the real disposable budget income of the households. It is also expected that the size that this phenomenon has reached in recent years will show no sign of declining in the medium-long term in many European Countries, and it will probably cause a severe crisis in the welfare system and in the real estate market.

Despite real estate investments being closely linked to the urban, regulatory and economic contexts in which they are applied, it is possible to recognise synergies and shared features in defining elements of this housing crisis across the European Union (EU) member states, namely:

- the desire and need to provide affordable housing through the construction and lease of homes (Crook and Kemp 2014, Whitehead et al. 2012, Oxley 2012, Haffner and Heylen 2011);
- the definition of target groups either in socio-economic terms or in relation to other kinds of vulnerability;
- the pursuit of housing quality by achieving energy efficiency standards and reducing social exclusion (Czischke and Pittini 2007).

Since the demand for social/temporary housing is increasing and it is rapidly changing all over Europe due to the recent global recession, a series of problems need to be solved to streamline and modernize policies and procedures. These are:

- Managing the growing and complex demand of low-cost housing;
- Managing the limited financial resources to be allocated in SH initiatives;

- Tackling the lack of an interconnection between the citizens and the Public Sector (PS).

To address this social challenge, it seems necessary to overcome the logic of conventional SH policies (Lami and Abastante 2017).

In this complex panorama, the methodological approach illustrated in section 2 has been applied to an Italian case study, as a possible approach to tackle the selection of SH projects, starting from the experience of the *Programma Housing* (Torino, Italy). The *Programma Housing* is an operating entity of the Italian Bank Foundation *Compagnia di San Paolo* active in the SH sector. One of its main activities is the selection process aimed at screening large funding requests from third bodies that submit innovative SH projects (from 2007, 6,150,000 euros have been allocated). The nature of the current process is complex and requires a detailed consideration of internal and external factors as well as a lot of decision criteria and alternatives. Moreover the projects submitted are characterized by a double identity:

i)  a *technical identity* related to the construction or redesign of the existing buildings to respond to the housing needs;

ii)  a *social identity* related to the social help needed by the people hosted in the SH projects. Indeed, one of the distinctive features of the SH is the presence of social help activities devoted to beneficiaries in order to integrate them in the society.

Due to the aforementioned intrinsic identities, each SH project is unique, making the selection process extremely delicate and difficult.

The activities conducted by the *Programma Housing* turned out to be interesting and challenging to apply the methodological approach developed in this study for several reasons: i) the huge amount of alternatives and decision criteria to be evaluated and compared; ii) the heterogeneous nature of quantitative and qualitative decision criteria; iii) the uniqueness of the SH projects; iv) the possibility to interface with the DMs involved in the actual selection process.

The methodological approach developed has been applied to the ranking process of the selection decision for the *Programma Housing*. The ranking process here presented involved eight interactions with the DMs, the first one in September 2015 and the last one in June 2016, making the process nine month long (Table 2).

Due to the confidential nature of the *Programma Housing* data, it took some time to start up the process with the DMs. As highlighted in Table 2, the first interaction aimed at illustrating the methodological approach in order to give the DMs all the needed information to decide whether they were interested or not in our research. The second interaction was devoted to solve confidentiality matters related to the SH projects so far financed from the *Programma Housing.* After the aforementioned discussions, the DMs showed to be interested in the methodological approach reported in this paper and therefore they gave us access to the SH projects data.

**Table 2.** The main steps of the ranking process

| DATE | NATURE | SCOPE | EXPERTS INVOLVED |
|---|---|---|---|
| *Preparation process meetings* | | | |
| September 2015 | Technical and social | Discussion with the decision makers intended to illustrate the methodological approach | Architects, psychologists, sociologists, experts in SH sector and decision-making |
| January 2016 | Technical and social | Discussion with the decision makers aimed at defining the privacy level of the data | Architects, psychologists, sociologists, experts in SH sector and decision-making |
| *Decision process meetings* | | | |
| February 2016 | Technical and Social | Definition of the alternatives to be compared and construction of the set of considered SH projects | Architects, psychologists, sociologists, experts in SH sector and decision-making |
| March 2016 | Technical and Social | Definition of the decision criteria and checking of the projects' performances | Architects, psychologists, sociologists, experts in SH sector and decision-making |
| March 2016 | Technical | Definition of the reference levels and AHP pairwise comparisons of the technical decision criteria | Architects, experts in SH sector and decision-making |
| April 2016 | Social | Definition of the reference levels and AHP pairwise comparisons of the social decision criteria | Psychologists, sociologists, experts in SH sector and decision-making |
| May 2016 | Technical and Social | Definition of the preference information about the interaction between criteria and definition of the preference orders on the technical and social decision criteria | Architects, psychologists, sociologists, experts in SH sector and decision-making |
| June 2016 | Technical and Social | Definition of the preferences orders of the decision criteria and some alternative SH projects | Architects, psychologists, sociologists, experts in SH sector and decision-making |

It is important to underline that the ranking process reported in the next sections constitutes a methodological application based on SH projects already financed by the *Programma Housing.* Nevertheless, thanks to the proved availability of the DMs, we aim at applying it during a future actual selection process.

*3.1. Structuring of the decision process*

In order to start the ranking process, it was necessary to define the set of alternative SH projects to be considered as well as the decision criteria.

According to the DMs suggestion acquired, we decided to apply the methodological approach to 21 SH projects located in the Piedmont Region (Italy) and financed by the *Programma Housing* between 2012 and 2014. The reasons for choosing those projects among others are mainly related to the presence of both technical and social aspects and the availability of homogeneous information. In this sense, the SH projects are heterogeneous but comparable at the same time.

The following steps required an interaction with the DMs in order to define the decision criteria. It is interesting to stress out that the DMs rethought to their own procedure, starting from the actual selection process conducted by the *Programma Housing* and involving 18 decision criteria. As a matter of fact, the DMs decided to aggregate and reduce the decision criteria from 18 to 10 since they realized that some criteria were redundant. The decision criteria are described in Table in which the column "Max Score" reports the maximum score that each project can reach in terms of each criterion performance.

**Table 3.** Description of the decision criteria

| CRITERIA | DESCRIPTION | | MAX SCORE |
|---|---|---|---|
| Overall consistency (C1) | This criterion considers the overall consistency of the project's spaces. It comprises: the location of the SH project, the integration of different uses and the presence of shared rooms. | Maximize | 10 |
| Quality of the design project (C2) | This criterion considers different aspects of the design project. It comprises: the flexibility and modularity of the architecture, the accessibility of the building for disabled people and the energy performances of the building. | Maximize | 15 |
| Beds (C3) | Overall amount of beds provided by the SH project. | Maximize | 10 |
| Economic consistency (C4) | This criterion assesses the economic aspects of the design project. It comprises: the fairness of the parametric costs, the co-financing amount and the coherence of the economic contribution financed by the *Programma Housing* with respect to the SH project proposal. | Maximize | 15 |
| Euros/beds (C5) | Parametric costs calculated with respect to the number of beds. The considered costs are realization costs as well as the costs of the social help activities provided. | Minimize | 20,000 €/bed |
| Clarity and innovation (C6) | Clarity in the description of the project's objectives and coherence/innovation with respect to the actions planned. | Maximize | 15 |
| Human resources (C7) | Information about the amount the human resources, their roles and their costs provided by the SH project. | Minimize | 10 |
| Social tools and methodologies (C8) | Social tools, activities and methodologies adopted by the SH project in order to properly face the different disadvantage situations. | Maximize | 25 |

| Economic sustainability (C9) | Economic information about the long-term sustainability of the SH project. This economic criterion considers the overall management costs (technical, social and administrative) needed to support the SH project in a long-term period. | Maximize | 10 |
|---|---|---|---|
| Synergies (C10) | Potential partnerships and networks on the territory of intervention activated by the SH projects. | Maximize | 15 |

From Table it is possible to notice that the DMs decided to consider 5 criteria with social connotation and 5 criteria with technical connotation in order to balance the ranking process.

The set of considered SH projects and their evaluations with respect to the considered decision criteria are shown in Table 4. It is important to stress out that the performances' evaluations of the SH projects reported have been assigned by the DMs during the actual selection processes conducted in 2012 and 2014 (step 1 of the proposed methodology). The performances' evaluations have been in turn converted and aggregated in order to consider the decision criteria described in Table 3.

In Table 4, the column "ID" identifies the SH projects considered through codes since it is not possible to report the projects' names due to privacy constrictions.

**Table 4.** Set of considered SH projects and evaluations

| ID | DECISION CRITERIA | | | | | | | | | |
|---|---|---|---|---|---|---|---|---|---|---|
| | C1 | C2 | C3 | C4 | C5 | C6 | C7 | C8 | C9 | C10 |
| P1 | 9 | 12 | 24 | 10 | 7,500 | 12 | 8 | 18 | 8 | 8 |
| P2 | 10 | 0 | 6 | 5 | 8,450 | 9 | 6 | 12 | 10 | 13 |
| P3 | 8 | 11 | 8 | 8 | 17,000 | 10 | 6 | 15 | 2 | 15 |
| P4 | 10 | 8 | 20 | 10 | 2,900 | 13 | 8 | 10 | 8 | 10 |
| P5 | 8 | 6 | 6 | 8 | 17,500 | 10 | 6 | 20 | 2 | 5 |
| P6 | 5 | 1 | 8 | 6 | 9,500 | 14 | 10 | 25 | 8 | 10 |
| P7 | 10 | 6 | 5 | 4 | 3,260 | 12 | 8 | 17 | 7 | 5 |
| P8 | 10 | 8 | 10 | 7 | 7,500 | 10 | 6 | 12 | 3 | 13 |
| P9 | 7 | 4 | 20 | 9 | 4,750 | 11 | 8 | 13 | 3 | 9 |
| P10 | 8 | 8 | 21 | 7 | 6,667 | 11 | 10 | 15 | 0 | 14 |
| P11 | 9 | 8 | 8 | 8 | 12,500 | 15 | 9 | 23 | 5 | 10 |
| P12 | 10 | 5 | 8 | 9 | 20,000 | 1 | 2 | 2 | 0 | 4 |
| P13 | 10 | 13 | 15 | 8 | 8,000 | 9 | 5 | 14 | 10 | 11 |
| P14 | 10 | 4 | 8 | 7 | 15,000 | 7 | 6 | 12 | 6 | 6 |
| P15 | 8 | 5 | 7 | 6 | 8,714 | 14 | 8 | 21 | 2 | 12 |
| P16 | 8 | 11 | 8 | 9 | 12,500 | 7 | 6 | 15 | 0 | 10 |
| P17 | 7 | 4 | 24 | 7 | 5,000 | 6 | 7 | 15 | 6 | 10 |
| P18 | 7 | 2 | 4 | 7 | 13,750 | 13 | 10 | 22 | 0 | 1 |
| P19 | 9 | 14 | 23 | 10 | 6,957 | 9 | 5 | 14 | 6 | 6 |
| P20 | 8 | 8 | 23 | 6 | 7,609 | 6 | 2 | 13 | 5 | 8 |

| P21 | 5 | 7 | 15 | 5 | 4,000 | 6 | 3 | 5 | 2 | 3 |
|-----|---|---|----|----|-------|---|---|---|---|---|

## 3.2. Application of the proposed methodological approach

After structuring the ranking process and having all the needed information, we applied the proposed methodological approach described in Section 2.4.

First of all, in order to use the Choquet integral preference model and, consequently, the NAROR, the evaluations of each project with respect to the decision criteria need to be expressed on a common scale.

This is possible by using the AHP (Saaty 1980, Saaty 1990) but it would require 210 pairwise comparisons for each one of the 9 criteria without a pre-existing numerical evaluation considered criteria leading to a total of 1,890 pairwise comparisons.

In order to reduce the cognitive effort of the DMs, we proposed them to apply the AHP only to a set of reference evaluations on the scale of each criterion and to determine the normalized value of the evaluations on the 21 SH projects considered by using the procedure described in Section 2.4.

Therefore, we carried out two different discussions with the DMs to define the set of reference evaluations (step 2 of the methodology) and we asked them to compare the values shown in Table 5 (step 3 of the methodology). It should be mentioned that the definition of the reference evaluations could be fixed with a non "standardized" procedure taylor-made for each criterion of the decision problem. In our case this developed an interesting discussion among DMs: approaching their choices in this way, they were forced to rethink the entire evaluation process and/or to clarify some steps that are often intuitively conducted.

**Table 5.** Reference evaluations for the considered criteria

| C1 | C2 | C3 | C4 | C5 | C6 | C7 | C8 | C9 | C10 |
|----|----|----|----|------|----|----|----|----|-----|
| 0 | 0 | 4 | 0 | 2,500 | 0 | 0 | 0 | 0 | 0 |
| 5 | 5 | 7 | 4 | 5,000 | 7 | 5 | 10 | 5 | 7 |
| 8 | 8 | 10 | 8 | 10,000 | 11 | 7 | 20 | 7 | 11 |
| 10 | 10 | 20 | 10 | 15,000 | 15 | 10 | 25 | 10 | 15 |
|  | 15 | 25 |  | 20,000 |  |  |  |  |  |

As a consequence, the pairwise comparisons asked to the DMs were:

- 6 for the 4 reference evaluations of the criteria C1, C4, C6, C7, C8, C9 and C10;
- 10 for the 5 reference evaluations of the criterion C2, C3 and C5,

which gave a total of 72 pairwise comparisons. Two examples of the pairwise comparisons given by the DMs during the aforementioned discussions are reported in Table 6.

**Table 6.** Pairwise comparison matrices for the considered decision criteria

| C3 (BEDS) | | | | CR = 0,0277 | | C5 (EUROS/BEDS) | | | | CR = 0,0899 | |
|---|---|---|---|---|---|---|---|---|---|---|---|
| | 25 | 20 | 10 | 7 | 4 | | 2.500 | 5.000 | 10.000 | 15.000 | 20.000 |
| 25 | 1 | 1 | 3 | 7 | 9 | 2.500 | 1 | 3 | 5 | 7 | 8 |
| 20 | 1 | 1 | 3 | 7 | 9 | 5.000 | 1/3 | 1 | 4 | 6 | 7 |
| 10 | 1/3 | 1/3 | 1 | 3 | 7 | 10.000 | 1/5 | 1/4 | 1 | 5 | 6 |
| 7 | 1/7 | 1/7 | 1/3 | 1 | 3 | 15.000 | 1/7 | 1/6 | 1/5 | 1 | 2 |
| 4 | 1/9 | 1/9 | 1/7 | 1/3 | 1 | 20.000 | 1/8 | 1/7 | 1/6 | 1/2 | 1 |

Considering the normalized evaluation of the reference points obtained by the AHP (Table 7) and interpolating them as described in Section 2.4, we were able to obtain the normalized evaluations of the 21 SH projects considered with respect to all criteria reported in Table 8 (step 4 of the methodology).

**Table 7.** Reference evaluations for considered criteria and normalized values obtained by AHP

| C1 | Normalized | C2 | Normalized | C3 | Normalized | C4 | Normalized | C5 | Normalized |
|---|---|---|---|---|---|---|---|---|---|
| 0 | 0 | 0 | 0 | 4 | 0 | 0 | 0 | 2.500 | 1 |
| 5 | 0.2060 | 5 | 0.1165 | 7 | 0.0881 | 4 | 0.2505 | 5.000 | 0.5473 |
| 8 | 0.6398 | 8 | 0.4929 | 10 | 0.3664 | 8 | 0.6941 | 10.000 | 0.2314 |
| 10 | 1 | 10 | 0.8203 | 20 | 1 | 10 | 1 | 15.000 | 0.0317 |
| - | - | 15 | 1 | 25 | 1 | - | - | 20.000 | 0 |
| C6 | Normalized | C7 | Normalized | C8 | Normalized | C9 | Normalized | C10 | Normalized |
| 0 | 0 | 0 | 0 | 0 | 0 | 0 | 0 | 0 | 0 |
| 7 | 0.1807 | 5 | 0.1852 | 10 | 0.1111 | 5 | 0.1618 | 7 | 0.1202 |
| 11 | 0.4630 | 7 | 0.1516 | 20 | 0.5591 | 7 | 0.6143 | 11 | 0.5347 |
| 15 | 1 | 10 | 1 | 25 | 1 | 10 | 1 | 15 | 1 |

For example, to obtain the normalized value of the SH project P1 with respect to the criterion C5 (Euros/Beds), first of all we observed that its evaluation (7,500 euro) is in the interval of the references for C5 whose extremes are 5,000 euro and 10,000 euro. Since the normalized evaluations of the two reference evaluations obtained by AHP are 0.5473 and 0.2314, applying the equation (1) we get the normalized evaluation of C5 for the SH project P1 as follows:

$$u(7.500) = u(5,000) + \frac{u(10,000) - u(5,000)}{(10,000) - (5,000)} (7,500 - 5,000) = 0.3894.$$

Figure 3 shows how the normalized values for the whole scales of all the considered criteria. As one can observe, the AHP was necessary to put all reference evaluations on the same scale. Indeed, the ten subfigures show that the preferences provided by the DMs are far from being linear.

**Figure 3.** Normalized values for the reference evaluations obtained by AHP

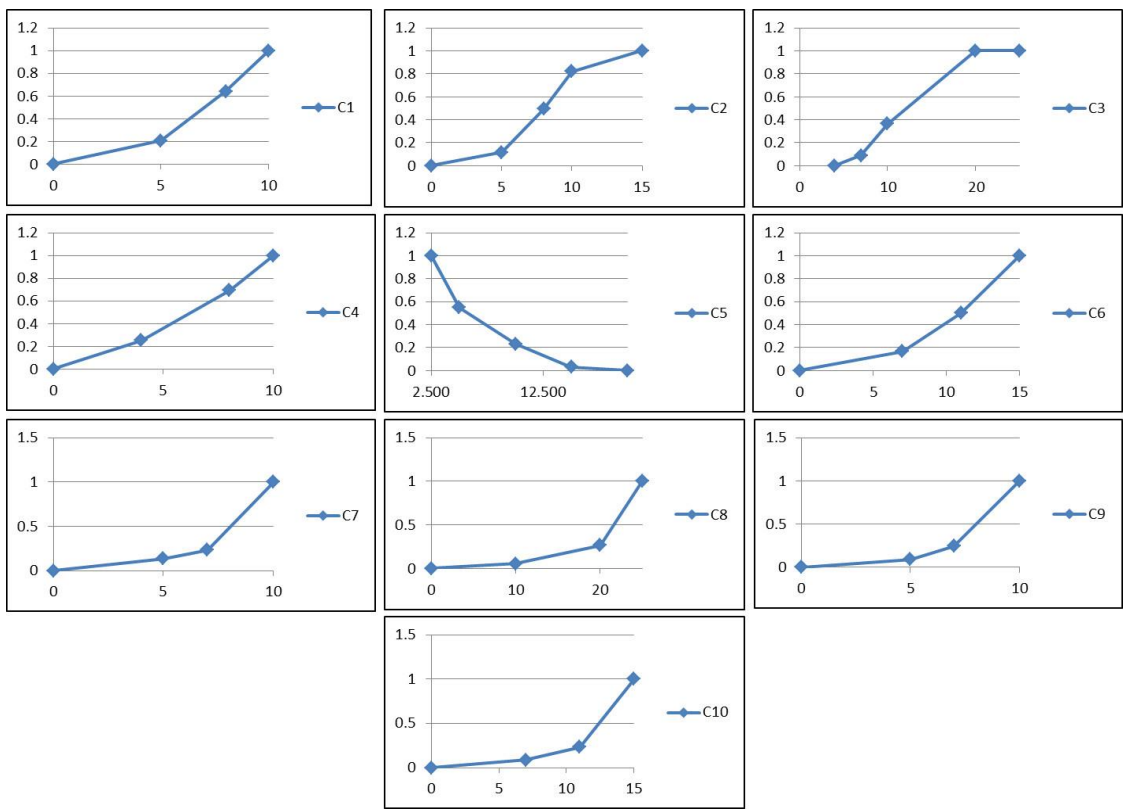

**Table 8.** Set of considered SH projects with normalized evaluations on each criterion

| ID | DECISION CRITERIA | | | | | | | | | |
|----|------|------|------|------|------|------|------|------|------|------|
| | C1 | C2 | C3 | C4 | C5 | C6 | C7 | C8 | C9 | C10 |
| P1 | 0.8199 | 0.8922 | 1 | 1 | 0.3894 | 0.5973 | 0.4344 | 0.4695 | 0.7429 | 0.2238 |
| P2 | 1 | 0 | 0.0587 | 0.3614 | 0.3293 | 0.3218 | 0.1684 | 0.2007 | 1 | 0.7674 |
| P3 | 0.6398 | 0.8563 | 0.1809 | 0.6941 | 0.0190 | 0.3924 | 0.1684 | 0.3351 | 0.0647 | 1 |
| P4 | 1 | 0.4929 | 1 | 1 | 0.9276 | 0.7315 | 0.4344 | 0.1111 | 0.7429 | 0.4311 |
| P5 | 0.6398 | 0.2420 | 0.0587 | 0.6941 | 0.0158 | 0.3924 | 0.1684 | 0.5591 | 0.0647 | 0.0858 |
| P6 | 0.2060 | 0.0233 | 0.1809 | 0.4723 | 0.2630 | 0.8658 | 1 | 1 | 0.7429 | 0.4311 |
| P7 | 1 | 0.2420 | 0.0294 | 0.2505 | 0.8624 | 0.5973 | 0.4344 | 0.4247 | 0.6143 | 0.0858 |
| P8 | 1 | 0.4929 | 0.3664 | 0.5832 | 0.3894 | 0.3924 | 0.1684 | 0.2007 | 0.0971 | 0.7674 |
| P9 | 0.4952 | 0.0932 | 1 | 0.8470 | 0.5926 | 0.4630 | 0.4344 | 0.2455 | 0.0971 | 0.3274 |
| P10 | 0.6398 | 0.4929 | 1 | 0.5832 | 0.4420 | 0.4630 | 1 | 0.3351 | 0 | 0.8837 |
| P11 | 0.8199 | 0.4929 | 0.1809 | 0.6941 | 0.1632 | 1 | 0.7172 | 0.8236 | 0.1618 | 0.4311 |
| P12 | 1 | 0.1165 | 0.1809 | 0.8470 | 0 | 0.0258 | 0.0741 | 0.0222 | 0 | 0.0687 |
| P13 | 1 | 0.9281 | 0.6832 | 0.6941 | 0.3578 | 0.3218 | 0.1852 | 0.2903 | 1 | 0.5347 |
| P14 | 1 | 0.0932 | 0.1809 | 0.5832 | 0.0317 | 0.1807 | 0.1684 | 0.2007 | 0.3880 | 0.1030 |
| P15 | 0.6398 | 0.1165 | 0.0881 | 0.4723 | 0.3126 | 0.8658 | 0.4344 | 0.6473 | 0.0647 | 0.6510 |
| P16 | 0.6398 | 0.8563 | 0.1809 | 0.8470 | 0.1315 | 0.1807 | 0.1684 | 0.3351 | 0 | 0.4311 |
| P17 | 0.4952 | 0.0932 | 1 | 0.5832 | 0.5473 | 0.1549 | 0.1516 | 0.3351 | 0.3880 | 0.4311 |
| P18 | 0.4952 | 0.0466 | 0 | 0.5832 | 0.0816 | 0.7315 | 1 | 0.7355 | 0 | 0.0172 |
| P19 | 0.8199 | 0.9641 | 1 | 1 | 0.4237 | 0.3218 | 0.1852 | 0.2903 | 0.3880 | 0.1030 |
| P20 | 0.6398 | 0.4929 | 1 | 0.4723 | 0.3825 | 0.1549 | 0.0741 | 0.2455 | 0.1618 | 0.2238 |
| P21 | 0.2060 | 0.3674 | 0.6832 | 0.3614 | 0.7284 | 0.1549 | 0.1111 | 0.0556 | 0.0647 | 0.0515 |

*3.2.1. Definition of the interactions between considered criteria*

To take into account the interaction between criteria, we applied the Choquet integral preference model. Therefore, considering the indirect preference information, a further meeting with the DMs was needed. During this meeting the DMs defined the preference information about the interaction between criteria. The 9 interactions provided are the:

1. Human resources (C7) and Synergies (C10) are positively interacting ($\varphi$(C7,C10) ≥ $\varepsilon$).

The DMs affirmed that this interaction reflects the ability of the SH projects to activate synergies and networks with the human resources already operating on the territory. In this sense, the project reduces the economic costs of the human resources.

2. Clarity and innovation (C6) and Economic sustainability (C9) are positively interacting $(\varphi(C6,C9) \geq \varepsilon)$.

According to the DMs suggestions, a well-structured and innovative SH project has the ability to attract private finances and it is therefore economically sustainable in a long-term period.

3. Beds (C3) and Euro/Beds (C5) are positively interacting $(\varphi(C3,C5) \geq \varepsilon)$.

The DMs affirmed that a valuable SH project should contain the costs in terms of euro/beds but, at the same time, increase the number of beds provided.

4. Beds (C3) and Economic sustainability (C9) are positively interacting $(\varphi(C3,C9) \geq \varepsilon)$.

The logical consequence of projects having a low number of beds is the low income coming from rents. As a matter of fact, a SH project is valuable for the DMs if it can ensure an economic sustainability in long-term period even in such circumstances.

5. Human resources (C7) and Economic sustainability (C9) are positively interacting $(\varphi(C7,C9) \geq \varepsilon)$.

In general, the economic sustainability of the SH projects requires an intricate administration, which needs in turn a huge amount of human resources.

The interaction here described reflects the ability of some SH projects of ensure an economic sustainability in long-term period having few human resources.

6. Beds (C3) and Human resources (C7) are positively interacting $(\varphi(C3,C7) \geq \varepsilon)$.

The DMs affirmed that this is a fundamental interaction because it contains the core of the SH concept; a sensible number of beds together with low human resources for the social activities means that the beneficiaries are autonomous as requested by the SH philosophy.

7. Overall consistency (C1) and Clarity and innovation (C6) are positively interacting $(\varphi(C1,C6) \geq \varepsilon)$.

Following the DMs reasoning, a clear and innovative SH project can only work if the location of the buildings, the internal/external spaces and the shared rooms are suitable to receive innovative social activities. Indeed, the SH projects are usually destined for people facing physical or psychological fragilities and requiring for specific social activities and spaces. In this sense, the *Programma Housing* pays attention to the concrete correspondence among the technical characters of the spaces and the social activities expected.

8. Beds (C3) and Economic consistency (C4) are positively interacting ($\varphi$(C3,C4) $\geq \varepsilon$).

   The DMs affirmed that providing a huge amount of beds as well as containing the economic consistency of the SH projects is usually a difficult task. Therefore, they decided to attribute a bonus to the projects reaching this objective.

9. Clarity and innovation (C6) and Human resources (C7) are positively interacting ($\varphi$(C6,C7) $\geq \varepsilon$).

   The DMs expressed their interest in innovative SH projects having a low amount of human resources. Usually, if the social activities planned in a SH project are very innovative, they could host also heterogeneous beneficiaries as, for example, young couples and elderly people. This means that people who needs for social help and people who don't, could help each other. In this sense, the human resources needed to manage the social activities are reduced and the related costs are cut down.

*3.2.2. Definition of the first rankings of decision criteria and alternative SH projects*

During the same meeting, the DMs were also asked to provide an order on the decision criteria in terms of their importance. After a long discussion, they decided to provide two different orders according to the social and technical criteria. The preference orders are the following:

- Importance order of social decision criteria: C8$\succ$ C7$\succ$ C6$\succ$ C9$\succ$ C10;
- Importance order of technical decision criteria: C1$\succ$ C4$\succ$ C2$\succ$ C5$\succ$ C3.

According to the interactions between criteria and to the order in terms of importance of social and technical criteria so far provided, we were able to present the DMs a first comprehensive order of the decision criteria according to their importance as measured by the Shapley value on the capacity obtained  computing the most representative value function  (Table 9).

**Table 9.** First ranking of the decision criteria

| CRITERIA RANKING | Shapley index |
|---|---|
| Social tools and methodologies (C8) | 0.1809 |
| Overall consistency (C1) | 0.1648 |
| Economic consistency (C4) | 0.1451 |
| Quality of the design project (C2) | 0.1253 |
| Human resources (C7) | 0.1009 |
| Clarity and innovation (C6) | 0.0812 |
| Economic sustainability (C9) | 0.0615 |
| Euros/beds (C5) | 0.0592 |
| Synergies (C10) | 0.0417 |
| Beds (C3) | 0.0395 |

A first ranking of the alternative SH projects, obtained by the most representative value function has been also presented to the DMs (Table 10).

**Table 10.** First ranking of the alternative SH projects

| POSITION | ID | Choquet Value | POSITION | ID | Choquet Value |
|---|---|---|---|---|---|
| 1° | P11 | 0.5733 | 12° | P7 | 0.3115 |
| 2° | P1 | 0.5311 | 13° | P8 | 0.3097 |
| 3° | P19 | 0.4617 | 14° | P9 | 0.2849 |
| 4° | P10 | 0.4240 | 15° | P5 | 0.2643 |
| 5° | P13 | 0.4230 | 16° | P20 | 0.2513 |
| 6° | P3 | 0.4152 | 17° | P17 | 0.2213 |
| 7° | P18 | 0.3955 | 18° | P2 | 0.2195 |
| 8° | P16 | 0.3955 | 19° | P21 | 0.2120 |
| 9° | P6 | 0.3955 | 20° | P14 | 0.1679 |
| 10° | P4 | 0.3955 | 21° | P12 | 0.1276 |
| 11° | P15 | 0.3559 | | | |

After seeing the two rankings (Tables 9 and 10), the DMs decided to add some information so far provided. The DMs stated that there were sensible differences among their thoughts and the rankings provided, according to their knowledge about the SH projects.

*3.2.2. Definition of the final rankings of decision criteria and alternative SH projects*

Despite the doubts of the DMs about the two rankings, at this phase of the process they were able to provide further information useful to come to more accurate results, as:

- An overall ranking of the decision criteria according to their importance expressed as follows:

C8≻ C1≻ C7≻ C6≻ C4≻ C2≻ C9≻ C5≻ C10≻ C3;

- A preference order on some SH projects that they were informed about. After a long internal discussion, the DMs arrived at the following preference order:

P1≻ P4≻ P10≻ P19≻ P6≻ P11.

Moreover, the DMs agreed on the fact that the 6 SH projects above are preferred to the other 15 alternative projects.

Denoting by $P$ the set composed of all different SH projects, this preference information is translated to the following linear inequalities:

- $C_\mu(\text{P1}) \geq C_\mu(\text{P4}) + \varepsilon$
- $C_\mu(\text{P4}) \geq C_\mu(\text{P10}) + \varepsilon$
- $C_\mu(\text{P10}) \geq C_\mu(\text{P19}) + \varepsilon$
- $C_\mu(\text{P19}) \geq C_\mu(\text{P6}) + \varepsilon$
- $C_\mu(\text{P6}) \geq C_\mu(\text{P11}) + \varepsilon$
- $C_\mu(\text{P11}) \geq C_\mu(Px)$ for all $Px \in P \setminus \{\text{P1, P4, P6, P10, P11, P19}\}$

Considering the preference information about the interaction between criteria (provided in Section 3.2.1) together with the overall ranking of the decision criteria and the preference SH project previously described, we were able to show to the DMs further results.

A final ranking of the decision criteria according to their importance as measured by the Shapley value has been calculated (Table 11).

**Table 11.** Final ranking of the decision criteria

| CRITERIA RANKING | Shapley index |
|---|---|
| Social tools and methodologies (C8) | 0.2050 |
| Overall consistency (C1) | 0.1250 |
| Human resources (C7) | 0.1200 |
| Clarity and innovation (C6) | 0.1150 |
| Economic consistency (C4) | 0.1100 |
| Quality of the design project (C2) | 0.1050 |
| Economic sustainability (C9) | 0.1000 |
| Euros/beds (C5) | 0.0950 |
| Synergies (C10) | 0.0150 |
| Beds (C3) | 0.0100 |

A final ranking of the alternative SH projects has been provided according to the Choquet integral corresponding to the most representative value function of each project.

**Table 12.** Final ranking of the alternative SH projects

| POSITION | ID | Choquet Value | POSITION | ID | Choquet Value |
|---|---|---|---|---|---|
| 1° | P1 | 0.5516 | 12° | P9 | 0.3180 |
| 2° | P4 | 0.5466 | 13° | P18 | 0.2999 |
| 3° | P10 | 0.4543 | 14° | P15 | 0.2939 |
| 4° | P19 | 0.4493 | 15° | P2 | 0.2647 |
| 5° | P6 | 0.4443 | 16° | P17 | 0.2647 |
| 6° | P11 | 0.4393 | 17° | P20 | 0.2647 |
| 7° | P13 | 0.4136 | 18° | P14 | 0.2613 |
| 8° | P7 | 0.3887 | 19° | P5 | 0.2584 |
| 9° | P8 | 0.3821 | 20° | P12 | 0.2398 |
| 10° | P3 | 0.3180 | 21° | P21 | 0.1692 |
| 11° | P16 | 0.3180 | | | |

## 4. Discussion of the results

It is possible to comment the results obtained by comparing the first and final rankings of the importance for the decision criteria (Table 9 and Table 11) and comprehensive evaluation of alternative SH projects (Table 10 and Table 12). It is important to remind that the first rankings were based on the preference information about interactions between criteria and two distinct importance ranking of social and technical criteria, while the final rankings were based on the interactions between criteria, the importance ranking of all criteria and the preference on some SH projects expressed by the DMs.

With respect to the importance rankings of decision criteria provided by the proposed methodological approach, we can state that few differences emerge.

**Table 13.** Main differences between the first and the final decision criteria rankings

| FIRST DECISION CRITERIA RANKING | | | FINAL DECISION CRITERIA RANKING | | |
|---|---|---|---|---|---|
| Position | Criterion | Shapley index | Position | Criterion | Shapley index |
| 1° | C8 | 0.1809 | 1° | C8 | 0.2050 |

| 2° | C1 | 0.1648 | 2° | C1 | 0.1250 |
| 3° | C4 | 0.1451 | 3° | C7 | 0.1200 |

Looking at the first three positions of both importance rankings of decision criteria (Table 13), outwardly we can say that the only difference is related to the third position occupied by the criterion C4 (Economic consistency) in the first ranking and the criterion C7 (Human resources) in the final one. However it is important to highlight the differences in terms of Shapley index. Indeed, in considering the two different criteria rankings, criterion C8 is the most important one but its difference from the second (C1) is lower than the difference between the same criteria in the final importance ranking of the criteria. In this second case, C8 is, without any doubt, the most important criterion while C1 and C7 are quite similar. In fact, according to the new information provided by the DMs, the importance of the criterion C8 (Social tools and methodologies) with respect to the criterion C1 (Overall consistency) substantially increases in the final decision criteria ranking (the difference of Shapley index between C8 and C1 is increased 4 times). Moreover, the Shapley index of the criterion C7 is almost equal to the Shapley index of the criterion C1.

It is interesting to notice that the criterion C8 is always considered the most important one even if it is never mentioned in the preference information about the interaction between criteria (Section 3.2.1) given by the DMs. This means that the performance of the SH projects on the criterion C8 are fundamental for the assessment process.

With respect to the alternative SH projects rankings provided, sensible differences can be noticed (Table 14).

**Table 14.** Main differences between the first and the final alternative SH projects rankings

| FIRST ALTERNATIVE RANKING | | | FINAL ALTERNATIVE RANKING | | |
|---|---|---|---|---|---|
| POSITION | ID | Choquet Value | POSITION | ID | Choquet Value |
| 1° | P11 | 0.5733 | 1° | P1 | 0.5516 |
| 2° | P1 | 0.5311 | 2° | P4 | 0.5466 |
| 3° | P19 | 0.4617 | 3° | P10 | 0.4543 |

A change in the preference orders between the first and the final alternative rankings is immediately recognisable. In fact, the only project appearing in the two rankings is P1, which obtains the second position in the first ranking and the first position in the final one.

## 4.1. First ranking analysis

In order to better understand the meaningfulness of the first ranking results reported in Table 14, the projects P11, P1, P19 have been analysed according to:

- the performance of each project (Table 8) with respect to the criteria C8, C1 and C4 (resulting as the most important ones in the first decision criteria ranking);
- the interaction between criteria given by the DMs and involving the three aforementioned criteria.

The project P11 (with a Choquet integral value of 0.5733) shows very good performances on criterion C8 (0.8236) and C1 (0.8199) and good performances on criterion C4 (0.6941). Moreover, as stated by the DM, C1 and C6 are positively interacting and the performances of P11 on these two criteria are excellent (0.8199 and 1, respectively). In line with this analysis, it seems therefore justified to see P11 in the first position of the first ranking.

The project P1 (with a Choquet integral value of 0.5311) has low performances on criterion C8 (0.4695) but very good performances on C1 (0.8199) and excellent on C4 (1). Because of the positive interaction between C3 and C4 stated from the DM and the good performances of P1 on these two criteria (both of them equal to 1), its second position can be partly justified by the low performance on C8.

Analogously, the project P19 (with a Choquet integral value of 0.4617) has very low performances about C8 (0.2903). This permits it to attain the third position of the first ranking. In fact, the performances of this project on criteria C1 and C4 are identical to that ones of project P1.

## 4.2. Final ranking analysis

In the same way, we explored the final ranking results reported in Table 14. The projects P1, P4 and P10 have been analysed according to:

○ the performance of each project (Table 8) with respect to criteria C8, C1 and C7 (resulting as the most important ones in the final decision criteria ranking);

○ the interaction between criteria given by the DMs and involving the three aforementioned criteria.

The performances of the project P1 (with a Choquet integral value of 0.5516) with respect to criteria C8 and C1 are identical to the ones mentioned for the first ranking. The performances of this project on criterion C7 (0.4344) are very low. However, due to positive interactions of C7 with the criteria C9, C3 and C6 expressed from the DM, the project P1 gets the first position of the final ranking. It is important to highlight that this project has been also specified as the most important one by the DMs in the preference order on some SH projects that they were informed about.

The project P4 (with a Choquet integral value of 0.5466) shows very low performances on C8 (0.1111) while that ones on C1 are excellent (1). Moreover, the positive interaction between C1 and C6 and its good performances on these two criteria (1 and 0.7315) give an added value to this project. It is important to notice that the overall performances of P1 and P4 are very similar in terms of the Choquet values. In case of P4, the preference information expressed in terms of ranking of some SH projects turned out to be fundamental for the position of this project in the final ranking. In fact, if we do not consider this information, the project P4 would result better than P1 according to the other information.

The project P10 (with a Choquet integral value of 0.4543) has low performances on C8 (0.3351), very good performances on C1 (0.6398) and excellent performances on C7 (1).

Starting from the analyses of the first and the final rankings it is possible to provide general reflections. First, criterion C8 (Social tools and methodologies) is confirmed being the most important because of its intrinsic nature. In fact, even if it does not show interactions with other criteria, good performances on criterion C8 are fundamental for a SH project to reach the top of the rankings. This is in line with the actual selection process adopted by the *Programma Housing* for which criterion C8 can contribute with a very high maximum score to the comprehensive evaluation of SH projects (Table 3). Moreover, this reflections agree with the SH basic idea considering the social tools and activities indispensable for the SH projects.

Second, criterion C7 (Human resources) showed to be fundamental in the second ranking due to its numerous interactions with other criteria, which are able to support the performances of C7.

Finally, the emerged differences between the first and the final rankings highlight the importance of the information provided by the DMs. In fact, the final rankings are based on different and more accurate information with particular reference to the preference order on some SH projects stated by the DMs. In this sense, the further information acquired for the final ranking has been fundamental to come to sensible and interesting results for the DMs.

## 5. Conclusions

In this paper we progressed both from the methodological and from the application point of view in Multiple Criteria Decision Aiding (MCDA). On one hand, a development of the Analytic Hierarchy Process (AHP) has been provided. It permits to use AHP also in case of decision problems presenting a high number of alternatives or criteria since it reduces the cognitive effort asked to the Decision Maker (DM) in a considerable way. The new proposal is composed of four different steps: (i) estimate of the performances of the considered alternatives on each criterion; (ii) definition of some reference evaluations for each considered criterion in accordance with the analyst; (iii) pairwise comparison of the reference evaluations by using the AHP method; (iv) interpolation of the values obtained by AHP in the previous step in order to get the normalized evaluations of all alternative performances. The new proposal has been tested on a group of undergraduate students belonging to the University of Catania and the test confirmed the goodness of the proposed methodology that permits to build a common scale of evaluations on all considered criteria.

On the other hand, the new proposal, that has been integrated with the Choquet integral and the Robust Ordinal Regression under a unified framework, has been applied to a real world problem being the evaluation of alternative Social Housing projects sited in the Piedmont region (Italy). Scholars have viewed that SH needs to be informed by increasingly sophisticated conceptions that treat the setting as a complex, multidimensional field, with social, economic and financial balance, environmental and quality of life issues. This is the realm, where the concepts get fuzzier, therefore requiring the use of new methods and theory that draws in new category to study. In this sense the application illustrated in the paper, which involved an interaction with different DMs that was nine months long, showed how the proposed method can help to have a better understanding of the problem at hand. The methodology proved to be useful in: i) stimulating the discussion with the stakeholders; ii) re-thinking about the decision criteria; iii) re -thinking about the reference levels.

The procedure conjugates the advantages of AHP in building a measurement scale and the advantages of the Choquet integral in handling interaction between criteria. In this context, the adoption of NAROR seems very beneficial because it permits to avoid focusing on only one capacity, which can be misleading for the reliability of the final decision.

We want also to underline that the proposed procedure permits to apply AHP also in decision problems with a large number of elements to be compared. Therefore, we believe that our proposal of a parsimonious version of AHP can be considered a relevant contribution for the basic theory and application of AHP.

**Acknowledgments**

*We would like to show our deep gratitude to the Programma Housing team who provided insight and expertise that greatly assisted the research, although they may not agree with all of the interpretations/conclusions of this paper*.

The second and the third authors wish to acknowledge the funding by the "FIR of the University of Catania BCAEA3, New developments in Multiple Criteria Decision Aiding (MCDA) and their application to territorial competitiveness".

## References

Angilella S, Corrente S, Greco S, Słowiński R. Robust Ordinal Regression and Stochastic Multiobjective Acceptability Analysis in multiple criteria hierarchy process for the Choquet integral preference model. *Omega* 2016; 63: 154-169.

Angilella S, Greco S, Matarazzo B. Non-additive robust ordinal regression: a multiple criteria decision model based on the Choquet integral. *European Journal of Operational Research* 2010a; 201(1): 277-288.

Angilella S, Greco S, Matarazzo B. The most representative utility function for non-additive robust ordinal regression. In: Hullermeier E, Kruse R, Hoffmann F, editors. *Proceedings of IPMU 2010*, LNAI 6178. Heidelberg: Springer; 2010b. p. 220-229.

Brans JP, Vincke P. A preference ranking organisation method: the PROMETHEE method for MCDM. *Management Science* 1985; 31(6): 647-656.

Choquet G. Theory of capacities. *Annales de l'institut Fourier* 1953; 5: 131-295.

Corrente S, Greco S, Ishizaka A. Combining analytical hierarchy process and Choquet integral within non-additive robust ordinal regression. *Omega* 2016; 61: 2-18.

Corrente S, Greco S, Kadziński M, Słowiński R. Robust ordinal regression in preference learning and ranking. *Machine Learning* 2013; 93: 381-422.

Crawford G, Williams C. A Note on the Analysis of Subjective Judgement Matrices. *Journal of Mathematical Psychology* 1985; 29(4): 387-405.

Crook T, Kemp PA. *Private Rental Housing*. Cheltenham: Edward Elgar; 2014.

Czischke D, Pittini A, *CECODHAS Housing Europe. Housing Europe 2007: Review of social, co-operative and public housing in the 27 EU member states*. CECODHAS; 2007.

Grabisch M. The application of fuzzy integrals in multicriteria decision making. *European Journal of Operational Research* 1996; 89: 445-456.

Greco S, Figueira JR, Ehrgott M. *Multiple Criteria Decision Analysis: State of the Art Surveys*. Berlin: Springer; 2016.

Greco S, Matarazzo B, Słowiński R. Rough set theory for multicriteria decision analysis. *European Journal of Operational Research* 2001; 129(1): 1-47.

Greco S, Mousseau V, Słowiński R. Ordinal regression revisited : multiple criteria ranking using a set of additive value functions. *European Journal of Operational Research* 2008; 191(2): 416-436.

Haffner M, Heylen K. User Costs and Housing Expenses. Towards a more Comprehensive Approach to Affordability. *Housing Studies* 2011; 26(4): 593-614.

Ishizaka A, Nemery, P. *Multicriteria decision analysis: methods and software*. Chichester, John Wiley & Sons, 2013.

Ishizaka A, Nguyen Nam H. Calibrated Fuzzy AHP for current bank account selection. *Expert Systems with Applications* 2013; 40(9): 3775–3783.

Keeney RL, Raiffa H. *Decisions with multiple objectives: preferences and value tradeoffs*. New York, Wiley; 1976.

Lami IM, Abastante F. Social Housing evaluation procedures: literature review and step forward. *Geoingegneria ambientale e mineraria* GEAM 2017; 149: (in press).

Marx I, Nolan B. *In-Work Poverty*. AIAS, GINI Discussion Paper 2012; 51 ( the paper can be downloaded from the website www.gini-research.org).

Meesariganda B R, Ishizaka A. Mapping verbal AHP scale to numerical scale for cloud computing strategy selection. *Applied Soft Computing* 2017, 53: 111–118.

Miller GA. The magical number seven, plus or minus two: some limits on our capacity for processing information. *Psychological review* 1956; 63(2): 81-97.

Millet I. The effectiveness of alternative preference elicitation methods in the Analytic Hierarchy Process. *Journal of Multi-Criteria Decision Analysis* 1997; 6(1): 41-51.

Mousseau V, Figueira JR, Dias L, Gomes da Silva C, Climaco J. Resolving inconsistencies among constraints on the parameters of an MCDA model. *European Journal of Operational Research* 2003; 147(1): 72-93.

Murofushi S, Soneda T. Techniques for reading fuzzy measures (III): interaction index. In: *9th Fuzzy Systems Symposium*, Sapporo, Japan; 1993. p.693–96.

Oxley M. Supply-side subsidies for affordable rental housing. In: Smith SJ, editor. *International Encyclopaedia of Housing and home*. Oxford: Elsevier Science; 2012.

Por HH, Budescu D. Eliciting Subjective Probabilities through Pair-wise Comparisons. *Journal of Behavioral Decision Making advance* online publications 2017. DOI: 10.1002/bdm.1929.

Rota GC. On the foundations of combinatorial theory I. Theory of Möbius functions. *Wahrscheinlichkeitstheorie und Verwandte Gebiete* 1964. p. 340-368.

Roy B. *Multicriteria methodology for decision aiding*. Dordrecht: Kluwer; 1996.

Saaty T. A scaling method for priorities in hierarchical structures. *Journal of Mathematical Psychology* 1977; 15(3): 234-281.

Saaty T. *The analytic hierarchy process*. New York: MacGraw-Hill; 1980.

Saaty T. How to make a decision: the analytic hierarchy process. *European Journal of Operational Research* 1990; 48(1): 9-26.

Saaty, T. and Ozdemir, M.S. Why the magic number seven plus or minus two. *Mathematical and Computer Modelling* 2003; *38*(3-4), pp.233-244.

Shafer G. *A mathematical theory of evidence*. Princeton NJ: Princeton University Press; 1976.

Shapley LS. A value for *n*-person games. In: Kuhn HW, Tucker AW, editors. *Contributions to the theory of games II*. Princeton: Princeton University Press; 1953. p. 307-317.

Sugeno M. *Theory of fuzzy integrals and its applications*. PhD thesis. Japan: Tokyo Institute of Technology; 1974.

Whitehead C, Monk S, Scanlon K, Markkanen S, Tang C. *The Private Rented Sector in the New Century—A Comparative Approach*. Cambridge: Cambridge Centre for Housing and Planning Research 2012: p. 27–38.

Wills J, Linneker B. In-work poverty and the living wage in the United Kingdom: a geographical perspective. *Transactions of the Institute of British Geographers* 2014; 39(2): 182-194.